\documentclass[10pt,twocolumn,letterpaper]{article}
\usepackage[british,english,american]{babel}

\usepackage[pagenumbers]{cvpr}

\usepackage{multirow}
\usepackage[table,xcdraw]{xcolor}
\usepackage{float}
\usepackage{microtype}
\usepackage{graphicx}
\usepackage{subcaption}
\usepackage{booktabs}
\usepackage{array}
\usepackage{enumitem}
\usepackage{hhline}
\usepackage{boldline}
\usepackage[figuresright]{rotating}
\usepackage{balance}
\usepackage{paralist}
\usepackage{arydshln}
\usepackage{wrapfig}
\usepackage[most]{tcolorbox}
\tcbuselibrary{listings,breakable}
\usepackage{fontawesome5}
\usepackage{xspace}
\usepackage{setspace}
\usepackage{relsize}
\usepackage{pifont}
\usepackage{amsmath}
\usepackage{amssymb}
\usepackage{mathtools}
\usepackage{amsthm}
\usepackage{bbm}
\usepackage{hyperref}

\definecolor{BlueGreen}{RGB}{0, 128, 128}
\definecolor{cvprblue}{rgb}{0.21,0.49,0.74}

\renewcommand{\paragraph}[1]{\vspace{1.25mm}\noindent\textbf{#1}}

\theoremstyle{plain}

\theoremstyle{definition}

\theoremstyle{remark}

\title{LoopViT: Scaling Visual ARC with Looped Transformers \vspace{-.4em}}

\author{Wen-Jie Shu$^{1}$ \quad Xuerui Qiu$^2$ \quad Rui-Jie Zhu$^3$ \quad Harold Haodong Chen$^1$ \quad Yexin Liu$^1$ \quad Harry Yang$^1$ \\[.3em]
$^1$HKUST \quad $^2$CASIA \quad $^3$UC Santa Cruz \\
 \texttt{wenjieshu2003@gmail.com}}

\begin{document}

\maketitle

\begin{abstract}
Recent advances in visual reasoning have leveraged vision transformers to tackle the ARC-AGI benchmark. However, we argue that the feed-forward architecture, where computational depth is strictly bound to parameter size, falls short of capturing the iterative, algorithmic nature of human induction. 
In this work, we propose a recursive architecture called \textbf{Loop-ViT}, which decouples reasoning depth from model capacity through weight-tied recurrence. 
Loop-ViT iterates a weight-tied \textit{Hybrid Block}, combining local convolutions and global attention, to form a latent chain of thought. 
Crucially, we introduce a parameter-free \textit{Dynamic Exit} mechanism based on predictive entropy: the model halts inference when its internal state ``crystallizes" into a low-uncertainty attractor. 
Empirical results on the ARC-AGI-1 benchmark validate this perspective: our 18M model achieves 65.8\% accuracy, outperforming massive 73M-parameter ensembles. These findings demonstrate that adaptive iterative computation offers a far more efficient scaling axis for visual reasoning than simply increasing network width. The code is available at \href{https://github.com/WenjieShu/LoopViT}{https://github.com/WenjieShu/LoopViT}.

\end{abstract}
    
\section{Introduction}
\label{sec:intro}

\begin{figure}[!t]
\centering
% \vspace{-0.4em}
\includegraphics[width=1.05\linewidth]{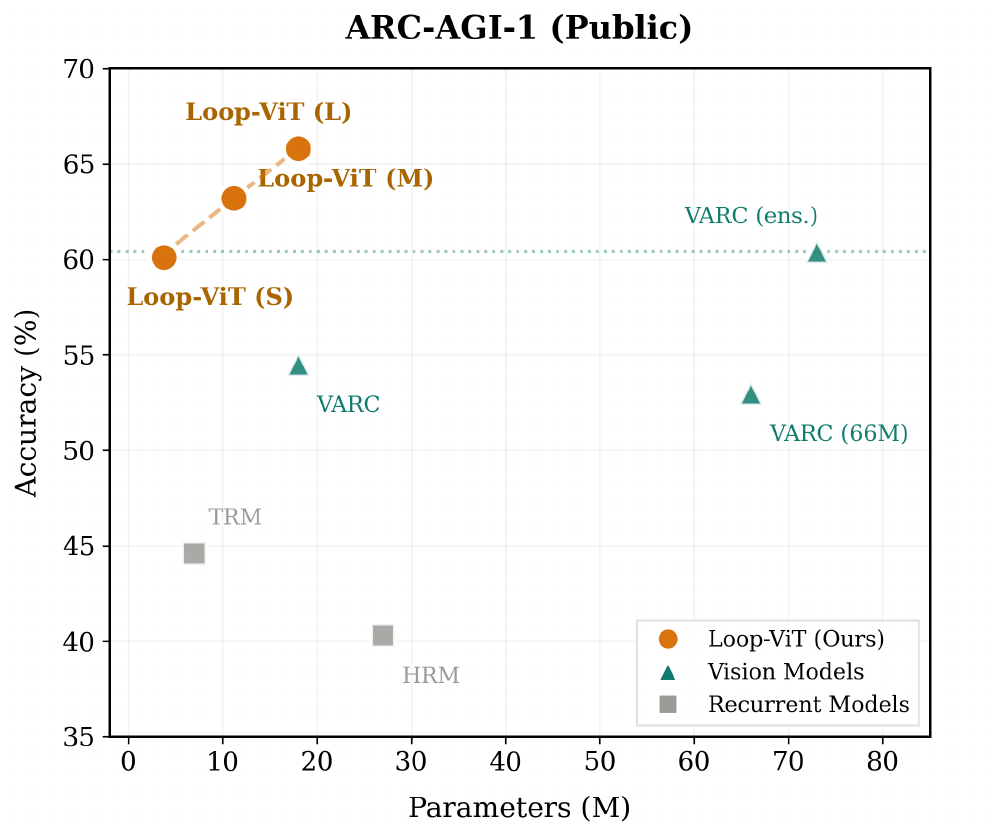}
\vspace{-2em}
\caption{Acc-Params comparisons with recurrent and vision methods. The vertical axis is accuracy (ARC-AGI-1); the horizontal axis is Parameters (memory cost). Our Loop-ViT outperforms previous methods while requiring significantly cheaper Params.
}
\label{fig:figure1}
\vspace{-1em}
\end{figure}

\begin{figure}[!t]
\centering
% \vspace{-0.4em}
\includegraphics[width=1.00\linewidth]{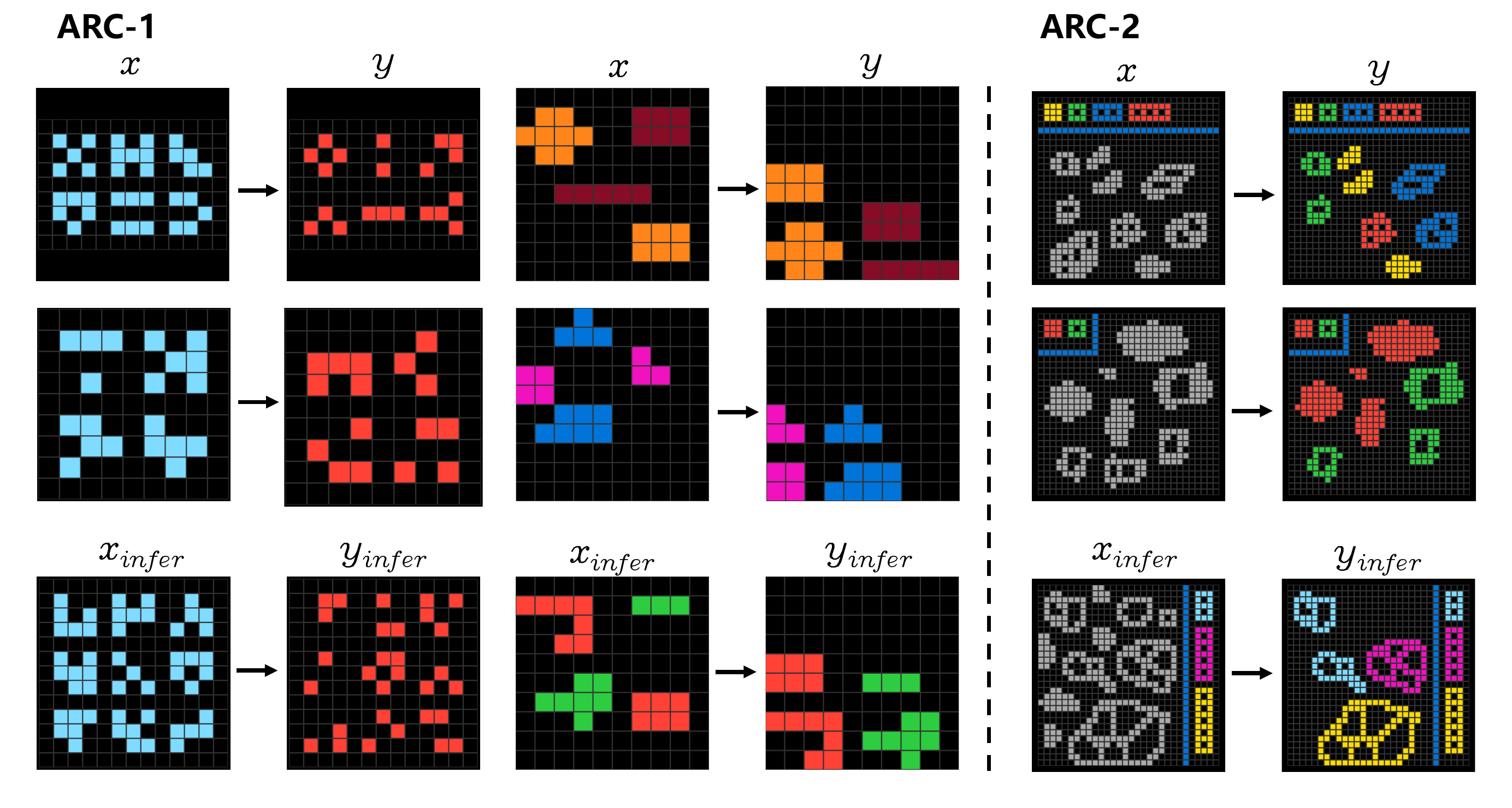}
\vspace{-2em}
\caption{\textbf{Illustration of ARC-AGI-1 and ARC-AGI-2 benchmarks.} The left two columns display tasks from ARC-AGI-1, characterized by visual priors such as ``Object Cohesion'' and ``Pattern Completion''. These tasks primarily test perceptual generalization. The right column showcases an ARC-AGI-2 task, exemplifying higher-order algorithmic challenges such as ``Symbolic Interpretation'', ``Compositional Reasoning'', and ``Contextual Rule Application''. For each task, the top rows show the few-shot demonstrations (Training) used to infer the rule, and the bottom row shows the query input (Inference).
}
\label{fig:figure2}
\vspace{-1em}
\end{figure}

A core facet of intelligence is \textit{visual reasoning}: inferring an underlying rule from a handful of examples and executing it in a novel setting. Importantly, the required reasoning depth varies widely across instances, ranging from simple rule discovery to multi-step execution. As illustrated in \cref{fig:figure2}, the Abstraction and Reasoning Corpus (ARC-AGI)~\cite{ARC, chollet2024arc} operationalizes this setting through visual grid puzzles that demand precise, compositional transformations (\textit{e.g.}, recursive filling, object relocation, or gravity-like dynamics) from only 2--4 demonstration pairs. Unlike conventional vision benchmarks that reward dataset-scale statistical learning of textures or semantics~\cite{imagenet, mscoco, cityscapes}, ARC emphasizes step-wise procedural reasoning. While humans solve these tasks through iterative hypothesis testing~\cite{johnson2021fast, legris2024h}, the benchmark remains challenging for deep learning systems that lack mechanisms for multi-step deliberation~\cite{pfister2025understanding, moskvichev2023conceptarc}.

Historically, ARC reasoning has relied on methods that serialize 2D grids into 1D sequences: program synthesis and Large Language Models (LLMs) convert grids to text  to exploit linguistic priors~\cite{wang2024hypothesis, berman2024record536, tang2024code, berman2024arcagi, li2025combining, berman2025highestscore, macfarlane2025searchinglatentprogramspaces}, while recurrent models~\cite{wang2025hierarchicalreasoningmodel, jolicoeurmartineau2025morerecursivereasoningtiny} process discrete grid tokens in a recurrent fashion. Both approaches, however, discard the spatial topology essential for visual reasoning. In contrast, the Vision ARC (VARC) framework~\cite{hu2025arc} demonstrated that vanilla Vision Transformers (ViTs)~\cite{dosovitskiy2021an} can solve ARC tasks directly from pixels, establishing that \textit{language is not necessary}: pure visual representations suffice for ARC-style visual reasoning.

Yet a key limitation persists: feed-forward ViTs scale inefficiently with reasoning complexity. As shown in \cref{fig:figure1}, simply increasing model capacity via depth or width yields diminishing returns, implying a structural mismatch for puzzles demanding recursion. We attribute this to the fact that visual reasoning is rarely a single-pass perceptual decision; it resembles an iterative latent deliberation process where an internal state is repeatedly updated. A feed-forward network, however, implements a fixed computation graph that forces a dynamic derivation into a static mapping. Our results (Sec.~\ref{sec:exp}) suggest that decoupling depth from parameter count via recurrence is a more effective scaling axis, allowing models to adapt computational effort (``Time'') rather than solely relying on raw capacity (``Space'').

To address this gap, we introduce \textbf{Loop-ViT}, a looped Vision Transformer tailored for pure visual reasoning. Loop-ViT replaces a stack of distinct Transformer layers with a weight-tied recurrent core executed for multiple iterations, decoupling computational depth from parameter count. This design encourages learning a reusable state-transition operator (a ``thought step'') rather than a collection of non-robust, task-specific heuristics. To better match the local, cellular-update nature of many ARC transformations, the recurrent core is implemented as a Hybrid Block that combines depth-wise convolutions with self-attention~\cite{shi2024transnextrobustfovealvisual, yu2022metaformeractuallyneedvision}. Finally, we introduce a Dynamic Exit mechanism driven by predictive entropy~\cite{shannon1948mathematical}: as predictions \textit{crystallize} (\textit{i.e.}, the output distribution stabilizes and entropy decays), Loop-ViT halts early on easier tasks, reducing average compute without compromising accuracy on hard reasoning problems.

Empirically, iterative computation proves a more efficient scaling axis than model capacity, as shown in \cref{fig:figure1}. Specifically: (I) \textbf{Pareto efficiency}: Loop-ViT improves the empirical Pareto frontier for visual reasoning when considering accuracy, compute, and parameters; (II) \textbf{Scalable performance}: a 3.8M parameter Loop-ViT (Small) reaches $60.1$\% score on ARC-1, surpassing the 18M VARC baseline ($54.5$\%) with roughly one-fifth of the parameters, and scaling the number of core layers further to 18M parameters (Large) improves to $65.8$\%, outperforming even large ensembles of feed-forward experts; (III) \textbf{Iterative refinement}: across iterations we observe consistent step-wise error decay and attention dynamics that shift from broad exploration to focused execution, suggesting an emergent deliberation process.

Our contributions are summarized as follows:
\noindent\textbullet~Introducing Looped Transformers to Vision: We propose Loop-ViT, the first looped Vision Transformer, establishing iterative recurrence as a powerful new paradigm for abstract visual reasoning.

\noindent\textbullet~A Performant and Efficient Design: Our architecture features a weight-tied Hybrid Block that aligns with recursive algorithms for robust reasoning, and a Dynamic Exit mechanism that enables adaptive ``thinking time" without extra parameters, significantly improving the accuracy-FLOPs trade-off.

\noindent\textbullet~Empirical Superiority over Parameter Scaling: We demonstrate that scaling through iteration is more effective than scaling parameters for abstract reasoning. Loop-ViT outperforms a larger state-of-the-art feed-forward model while using $\mathbf{5\times}$ fewer parameters.

\section{Related Work}
\label{sec:related_work}

\begin{figure}[!t]
\centering
% \vspace{-0.4em}
\includegraphics[width=1.00\linewidth]{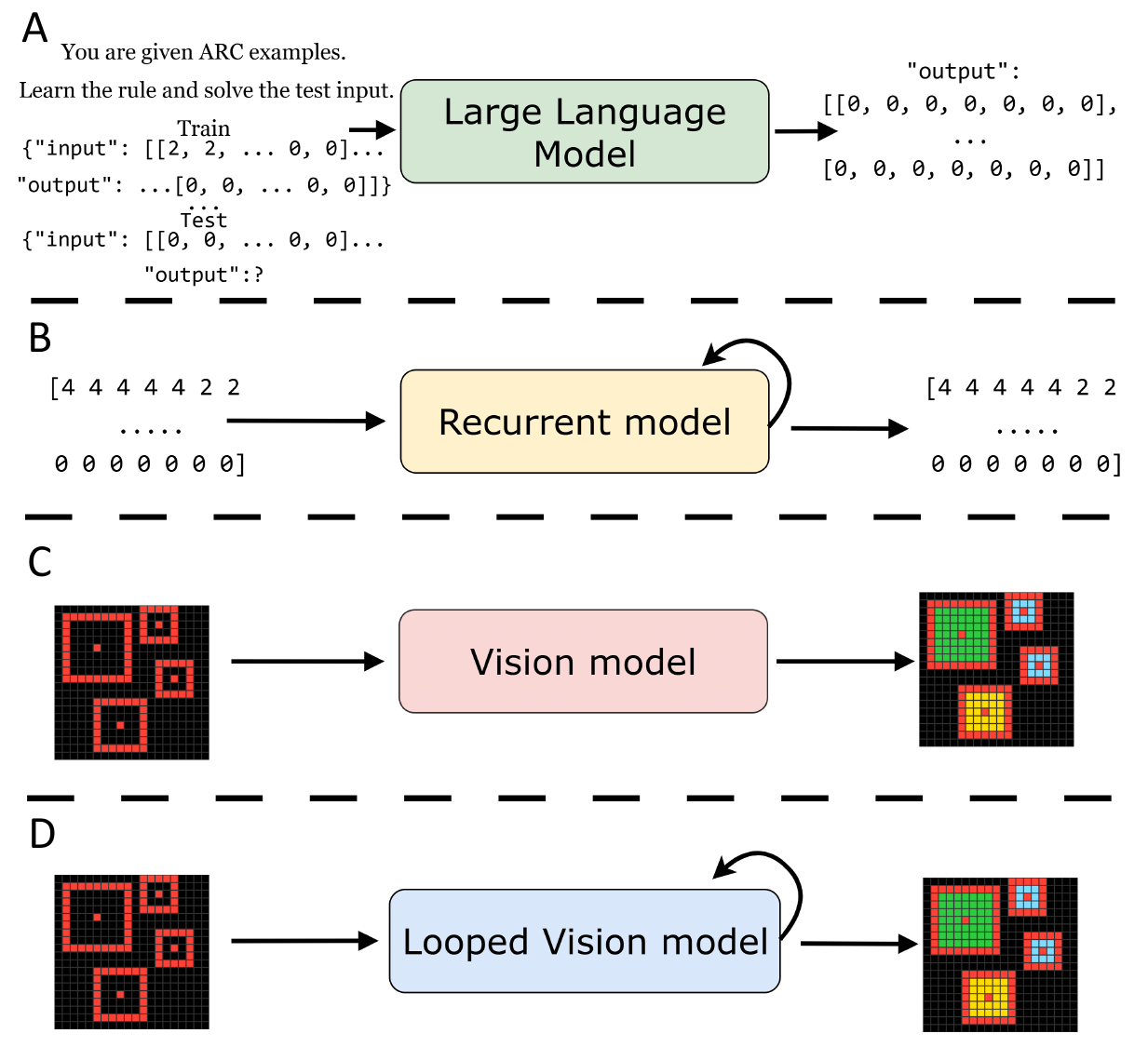}
\vspace{-2em}
    \caption{\textbf{Comparison of input representations and inference
paradigms for ARC.} (A) LLMs operate on a 1D textual token
sequence obtained by serializing the ARC grids into a prompt
(\textit{e.g.}, JSON/ASCII). (B) Recurrent token models also take a 1D
sequence, but with a discrete grid-tokenization that pads the grid to
a fixed canvas and inserts special boundary tokens (\textit{e.g.}, PAD/EOS),
yielding a fixed-length token stream. (C) VARC follows a vision
formulation, encoding the grid as a 2D spatial input processed
in a single forward pass. (D) Ours combines the vision input
with looped/iterative inference, repeatedly refining internal repre-
sentations and predictions across multiple steps, bridging spatial
inductive bias and recurrent computation.}
\label{fig:compare}
\vspace{-1em}
\end{figure}

We situate Loop-ViT within the broader landscape of visual reasoning, distinguishing it from language-centric approaches and clarifying the role of recurrence in modern architectures. \cref{fig:compare} visually summarizes the evolution of these paradigms.

\noindent\textbf{Paradigms of Abstract Reasoning: Language \textit{vs.} Vision.}\quad
Historically, ARC reasoning has relied on language-based methods (\cref{fig:compare} (A-B)), which serialize 2D grids into 1D sequences. This is typically done via JSON or ASCII representations for Large Language Models (LLMs)~\cite{wang2024hypothesis, berman2024arcagi}, or as discrete tokens for recurrent models~\cite{wang2025hierarchicalreasoningmodel, jolicoeurmartineau2025morerecursivereasoningtiny}. Although these methods exploit powerful linguistic priors, the serialization process inevitably discards the spatial topology essential for many visual puzzles.
Recently, the Vision ARC (VARC) framework~\cite{hu2025arc} shifted this paradigm to pure vision (\cref{fig:compare} (C)), treating reasoning as an image-to-image translation task. This formulation allows the use of powerful Vision Transformers (ViTs) and standard data augmentations. However, standard ViTs are feed-forward universal approximators and lack an inherent inductive bias for the \textit{iterative algorithm execution} required by complex ARC tasks. Our work retains the visual formulation of VARC but fundamentally alters the computational graph from a static pass to a dynamic loop (\cref{fig:compare} (D)).

\noindent\textbf{Looped Transformers and Algorithmic Generalization.}\quad
Reusing layer parameters across depth, often termed ``weight tying,'' allows neural networks to implement iterative algorithms instead of static pattern matching. In NLP, architectures like Universal Transformers~\cite{dehghani2018universal} and ALBERT~\cite{lan2019albert} established that such loops improve parameter efficiency and generalization. Recent work on ``thinking models''~\cite{banino2021pondernetlearningponder, liu2024looped, zhou2025looped, saunshi2025reasoning, zeng2025pretraining} further suggests that recurrence supports a latent Chain-of-Thought, enabling models to adapt their computation based on task complexity. Modern large-scale studies~\cite{geiping2025scaling, zhu2025scalinglatentreasoninglooped} indicate that scaling this ``thinking time'' in the latent space can achieve results comparable to much larger, deeper models.
In computer vision, recurrent processing has traditionally focused on refining continuous signals, such as optical flow estimation in RAFT~\cite{teed2020raft}. Earlier efforts also applied recurrent ResNets to synthetic maze-solving tasks~\cite{bansal2022endtoendalgorithmsynthesisrecurrent, schwarz2021can}, demonstrating the potential for algorithmic generalization. However, these methods often relied on convolutional priors tailored to specific, narrow domains. Loop-ViT extends this recurrent philosophy to modern Vision Transformers, treating weight-tied loops as a scalable primitive for abstract visual reasoning across diverse tasks.

\noindent\textbf{Adaptive Computation Mechanisms.}\quad
A unique advantage of recurrent architectures is the ability to decouple computation from parameter count. Classical ``soft halting'' approaches (ACT~\cite{graves2017adaptivecomputationtimerecurrent}, PonderNet~\cite{banino2021pondernetlearningponder}) treat halting as a probabilistic variable, requiring complex auxiliary losses. In contrast, ``hard halting'' strategies~\cite{xu2023lgvit, raposo2024mixture} rely on learned routing policies to adjust depth. Loop-ViT implies a simpler, parameter-free entropy exit. By monitoring the stabilization of predictive entropy (``crystallization''), our model halts when the internal state reaches a stable attractor, ensuring logical consistency without extra supervision or parameters.

\section{Method}
\label{sec:method}
\begin{figure*}[t] 
    \centering
    \includegraphics[width=1.00\linewidth]{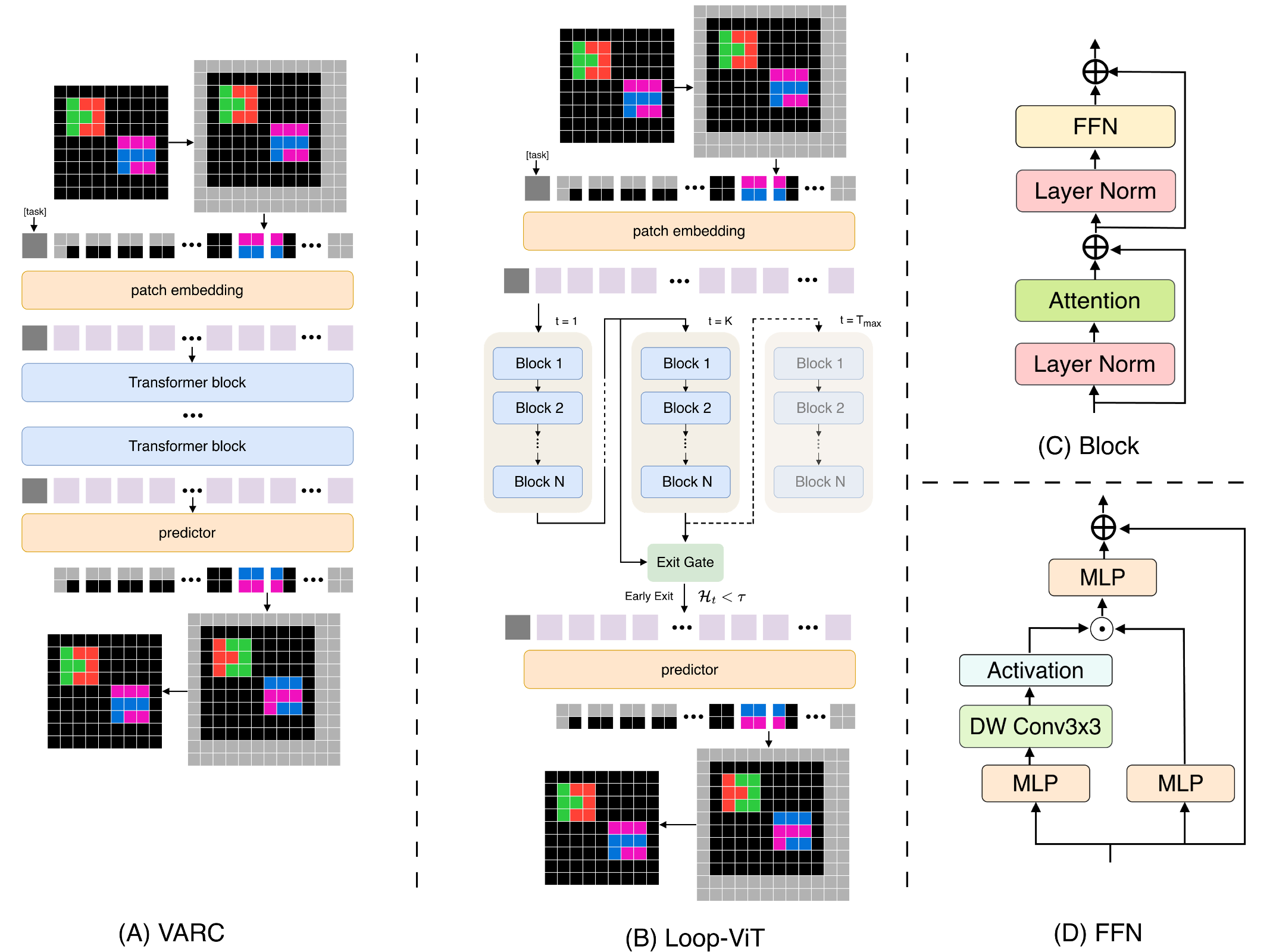} 
    \caption{\textbf{The overall pipeline of the proposed LoopViT.} (A) Comparison of the standard VARC pipeline versus our Loop-ViT pipeline. Loop-ViT introduces iterative state refinement through a weight-tied core. (B) Detailed unrolled view of the Loop-ViT recurrence, where the state $z_t$ acts as a dynamic memory. (C) Structure of the Hybrid Transformer Block, employing RMSNorm and Rotary Positional Embeddings. (D) The Heterogeneous Feed-Forward Network (ConvGLU), which splits processing pathways to apply depth-wise convolution solely to image tokens while preserving task tokens, reconciling local spatial updates with global rule induction.}
    \label{fig:overall_pipeline}
\end{figure*}

\cref{fig:overall_pipeline} illustrates the overall architecture of \textbf{Loop-ViT}. As depicted in \cref{fig:overall_pipeline} (B), our model introduces iterative state refinement through a weight-tied core. We first formalize this \textit{Global Recurrent Architecture} and the weight-tied layer reuse (Sec.~\S\ref{subsec:architecture}). We then detail the \textit{Hybrid Encoder Block}, which integrates convolutional and attention mechanisms for heterogeneous processing of image and task tokens (Sec.~\S\ref{subsec:hybrid_block}). Besides, we introduce the \textit{Dynamic Exit} strategy, which leverages predictive entropy to adaptively halt computation during inference. Finally, we describe the stable training protocol (Sec.~\S\ref{subsec:dynamic_exit})

\subsection{Loop-ViT Architecture}
\label{subsec:architecture}
Let $\text{emb}(\cdot) : \mathbb{X} \to \mathbb{R}^{M \times d}$ be the input embedding function that maps a visual grid $x$ and task-specific context $c$ to a sequence of $M$ tokens with hidden dimension $d$. Let $\mathcal{M}_\theta(\cdot) : \mathbb{R}^{M \times d} \to \mathbb{R}^{M \times d}$ be a core transformer trunk parameterized by $\theta$. As illustrated in \cref{fig:overall_pipeline} (A), a standard non-looped vision transformer stacks $L$ distinct layers such that the total function $\mathcal{F}$ is:
\begin{equation}
    \mathcal{F}(\cdot) := \text{head} \circ \mathcal{L}_{\theta_L} \circ \cdots \circ \mathcal{L}_{\theta_1} \circ \text{emb}(\cdot),
\end{equation}
where $\text{head}(\cdot)$ is the output projection layer. 
In contrast, our proposed \textbf{Loop-ViT} reuses the same core trunk $\mathcal{M}_\theta$ for $T$ iterations. Let $t \in \{1, \dots, T_{\max}\}$ be the number of loop steps. As shown in the unrolled view of \cref{fig:overall_pipeline} (B), the state $z_t$ evolves through the recursive application of the transition operator:
\begin{equation}
    z_{t+1} = \mathcal{M}_\theta(z_t + e_t), \quad z_0 = \text{emb}(\cdot),
\end{equation}
where $e_t$ is a learned step-dependent embedding that disambiguates the computation progress. The final output is given by $\mathcal{F}^{(t)}(\cdot) = \text{head}(z_t)$. This formulation forces the model to learn a unified, step-wise ``transition rule" that is robust enough to be applied repeatedly. Crucially, scaling the computational duration $T$ does not increase the parameter count, allowing the model to emulate complex algorithmic simulations with high efficiency. For inference iterations exceeding the training budget, we adopt an identity extrapolation for the step embeddings: $e_t = e_{T_{\text{train}}}$ for all $t > T_{\text{train}}$.
\subsection{Hybrid Encoder Block}
\label{subsec:hybrid_block}
We hypothesize that ARC tasks require two distinct modes of processing: \textit{local pattern matching} (e.g., continuing a line or filling a region) and \textit{global rule induction} (e.g., detecting symmetry or gravity). To support this, our \textbf{Hybrid Block} explicitly fuses the strengths of convolutions and attention. The depth-wise convolution in the FFN acts as a cellular automaton update rule, processing local neighborhoods to maintain spatial consistency. Simultaneously, the global attention mechanism broadcasts rule information across the entire grid, enabling long-range reasoning.

The core trunk $\mathcal{M}_\theta$ consists of $L$ hybrid encoder layers. Each layer balances global relational reasoning with local spatial updates.

\noindent\textbf{Internal Layer Structure.}\quad
As shown in \cref{fig:overall_pipeline} (C), each layer consists of Multi-Head Self-Attention (MHSA) followed by a Heterogeneous ConvGLU as the Feed-Forward Network (FFN). To better capture spatial relationships in ARC grids, the MHSA employs Rotary Positional Embeddings (RoPE) \cite{su2023roformerenhancedtransformerrotary}. Given an input sequence $Z \in \mathbb{R}^{M \times d}$, we first project it into query, key, and value manifolds for each head $h$:
\begin{equation}
    Q_h, K_h, V_h = Z W_h^Q, Z W_h^K, Z W_h^V.
\end{equation}
The RoPE operator $f_{\text{R}}$ is applied to $Q_h$ and $K_h$ to inject relative positional information. The output of a single head $O_h$ is then computed as:
\begin{equation}
    O_h = \text{Softmax}\left(\frac{f_{\text{R}}(Q_h) f_{\text{R}}(K_h)^T}{\sqrt{d_h}}\right) V_h,
\end{equation}
\\[0.5ex] % 增加一点行间距，避免下方衔接太紧
where $W_h^Q, W_h^K, W_h^V$ are learnable projections and $d_h$ is the head dimension. The final MHSA output integrates all heads via concatenation and a linear projection $W^O$:
\begin{equation}
    \text{MHSA}(Z) = \text{Concatenate}(O_1, \dots, O_H) W^O.
\end{equation}
To ensure numerical stability during deep recurrence, we adopt a pre-norm configuration with RMSNorm~\cite{zhang2019rootmeansquarelayer}. The full layer transition is defined as:
\begin{align}
    Z' &= Z + \text{MHSA}(\text{RMSNorm}(Z)) \\[1ex] % 这里增加行间距
    Z_{\text{out}} &= Z' + \text{ConvGLU}(\text{RMSNorm}(Z')).
\end{align}

\noindent\textbf{Heterogeneous Processing.}\quad
The primary engine of our visual induction is the Heterogeneous ConvGLU, as illustrated in \cref{fig:overall_pipeline} (D). Recognizing that task-level context tokens and spatial image patches require distinct inductive biases, we apply depthwise convolutions selectively. For a sequence $Z$, we first compute the gated and value representations:
\begin{equation}
    [X_{\text{gate}}, X_{\text{val}}] = \text{Linear}_1(Z).
\end{equation}
We then partition $X_{\text{gate}}$ into task tokens $G_{\text{task}}$ and image tokens $G_{\text{img}}$. While $G_{\text{task}}$ bypasses the spatial operator to preserve abstract rules, we reshape $G_{\text{img}}$ to a 2D grid and apply a $3 \times 3$ depthwise convolution (DW-Conv) to capture local connectivity:
\begin{equation}
    \hat{G}_{\text{img}} = \text{Flatten}(\text{DW-Conv}(\text{Reshape}(G_{\text{img}}))).
\end{equation}
The augmented gate $\hat{X}_{\text{gate}}$ is then reassembled via concatenation: $\hat{X}_{\text{gate}} = [G_{\text{task}}, \hat{G}_{\text{img}}]$. The final output is derived as:
\begin{equation}
    \text{ConvGLU}(Z) = \text{Linear}_2(\sigma(\hat{X}_{\text{gate}}) \odot X_{\text{val}}),
\end{equation}
where $\sigma$ denotes the activation function (e.g., SiLU). This dual-track prior facilitates the decomposition of visual reasoning: MHSA facilitates global task induction, while ConvGLU executes local spatial transformations.

\subsection{Dynamic Exit via Entropy-Based Prediction Crystallization}
\label{subsec:dynamic_exit}
A key insight of recurrent vision is that reasoning depth should be adaptive rather than fixed; ideally, computation should cease once a solution ``crystallizes." This design is motivated by the variable complexity of ARC tasks: while simple geometric transformations may stabilize in few iterations, complex algorithmic puzzles require prolonged refinement to resolve logical ambiguity.

To exploit this, we introduce an inference-time \textbf{Dynamic Exit} mechanism based on the magnitude of predictive entropy. Let $P_t = \text{softmax}(\text{head}(z_t))$ be the predicted probability distribution over the grid categories at step $t$. We quantify the model's confidence through the average pixel-wise Shannon entropy:
\begin{equation}
    \mathcal{H}_t = - \frac{1}{N} \sum_{i=1}^N \sum_{c=1}^C P_{t,i}(c) \log P_{t,i}(c),
\end{equation}
where $N$ is the number of pixels and $C$ is the number of color categories. During inference, generation halts at step $t$ if the entropy falls below a confidence threshold $\mathcal{H}_t < \tau$, where $\tau = 0.05$. If the threshold is not met, computation continues until reaching a hard limit of $T_{\max}$ iterations. Once halted, the state is ``frozen" ($z_{k} = z_t, \forall k > t$), effectively bypassing further core layer computations. This entropy-based strategy requires no additional parameters and provides a principled measure of when the model has reached a stable attractor in its latent space.

\subsection{Training Strategy}
\label{subsec:training}
We employ a training protocol designed to foster these stable recurrent dynamics.

\noindent\textbf{Fixed-Depth Training.}\quad
The model is trained on the ARC and RE-ARC datasets. Crucially, we do \textit{not} use dynamic halting during training. Instead, we unroll the core $\mathcal{M}_\theta$ for a fixed number of steps $T$ (e.g., $T=12$) and apply supervision on the final output. This procedure ensures that the model learns a robust transition rule $\mathcal{M}_\theta$ that converges to the correct solution within the allocated budget, rather than overfitting to early exits. The objective is a standard per-pixel cross-entropy loss:
\begin{equation}
    \mathcal{L}_{\text{offline}} = \text{CrossEntropy}(P_T, Y_{\text{gt}}).
\end{equation}

\noindent\textbf{Test-Time Training (TTT).}\quad
During evaluation, we fine-tune the shared weights $\theta$ on the few-shot demonstrations of each specific task. We generate augmented views (rotations, flips, color permutations) of the support examples to create a task-specific batch. This adaptation phase specializes the general-purpose ``thought step" into a dedicated algorithm for the current puzzle, further sharpening the convergence profile.
\section{Experiments}
\label{sec:exp}
This section evaluates \textbf{Loop-ViT} on the ARC-AGI. We test the hypothesis that visual reasoning is effectively modeled as a recurrent state transition rather than a fixed-depth feed-forward process. The evaluation is structured around three main findings: (\textbf{\textit{i}}) \textbf{Global Performance and Efficiency}, comparing Loop-ViT against LLMs and state-of-the-art vision baselines; (\textbf{\textit{ii}}) \textbf{Structural Scaling Laws}, exploring the interaction between space (parameters) and time (iterations); and (\textbf{\textit{iii}}) \textbf{Step-wise Attention Dynamics}, analyzing how internal attention patterns evolve across reasoning steps.

\subsection{Experimental Setup}
\label{subsec:setup}
\paragraph{Datasets and Benchmarks.}
Primary evaluation is conducted on the \mbox{ARC-AGI-1} benchmark~\cite{chollet2019measureintelligence}. Following state-of-the-art pixel-based methodologies~\cite{hu2025arc}, we augment the training split with synthetic samples from the RE-ARC generator~\cite{hodel2024addressingabstractionreasoningcorpus}. We report \textbf{Pass@2 accuracy} in percentage (\%). Results on the \mbox{ARC-AGI-2} set are also provided to assess out-of-distribution generalization.

\paragraph{Implementation Framework.}
A two-stage training pipeline is adopted: (i) offline pre-training on augmented datasets; (ii) Test-Time Training (TTT)~\cite{Sun2020TTT} on few-shot demonstrations. During TTT, the model specializes its weights to the specific task through local augmentations (\textit{e.g.}, rotations and flips).

\paragraph{Model Configurations.}
We analyze three variants designated as Small (3.8M params), Medium (11.2M params), and Large (18M params). We set $T_{\max}\!\in\![20, 28]$ for the Small variant and $T_{\max}\!\in\![4, 8]$ for Medium/Large, depending on specific model scale and task complexity.

\begin{figure*}[t] 
    \centering
    \includegraphics[width=1.00\linewidth]{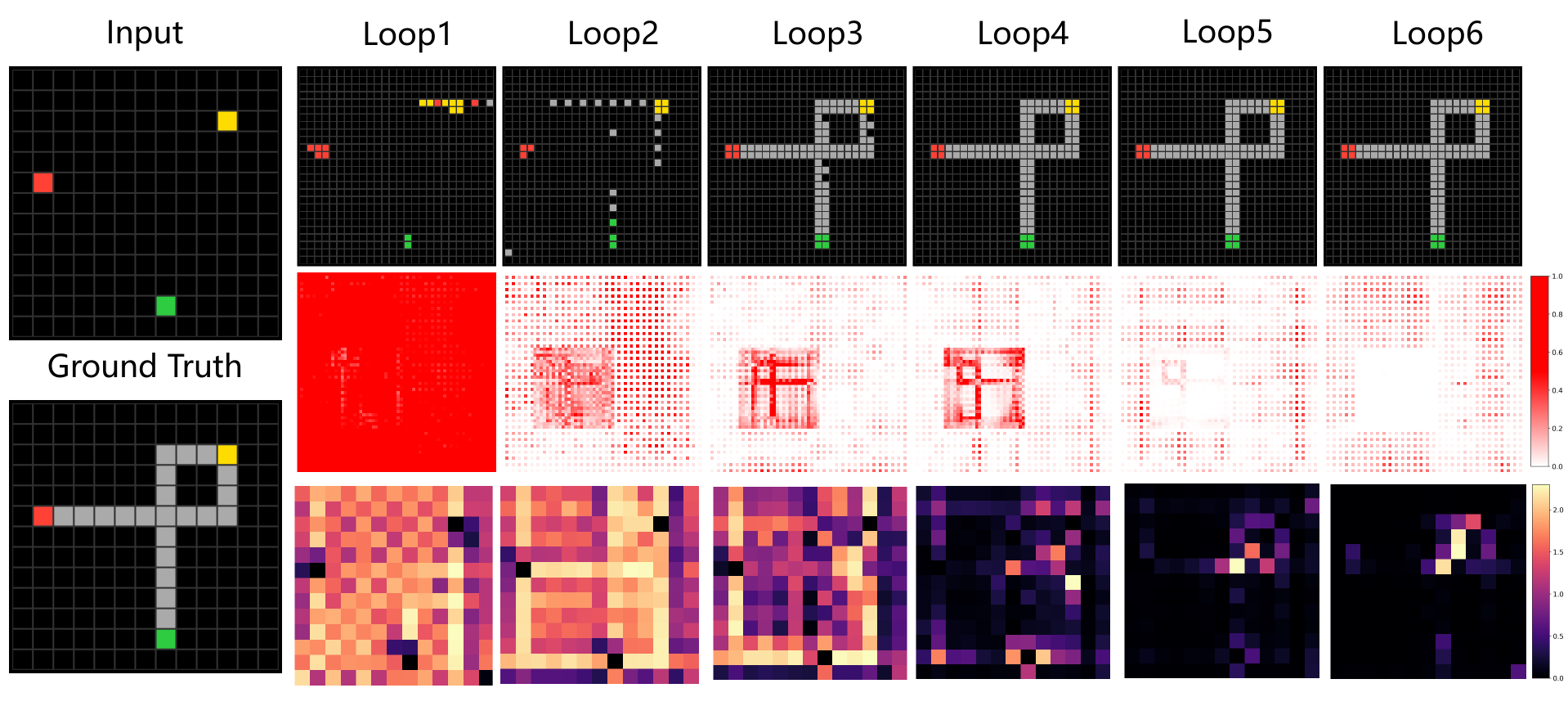} 
    \caption{\textbf{Iterative Prediction Refinement in Loop-ViT.} 
    (Top) The model's output progressively approaches the ground truth through successive iterations. 
    (Middle) Pixel-wise difference maps between consecutive steps show decreasing prediction volatility. 
    (Bottom) Entropy measurements demonstrate the stabilization of the model's confidence. 
    This ``crystallization effect" reveals how recurrent processing enables gradual convergence to logically consistent solutions.}
    \label{fig:overall}
\end{figure*}
\subsection{Pillar 1: Performance and Parameter Efficiency}
\label{subsec:main-results}
Table \ref{tab:arc_results} summarizes the performance of Loop-ViT. Our results confirm that dedicated visual reasoning architectures can achieve comparable or superior results to massive Large Language Models~\cite{guo2025deepseek, arcprize2025arcagi_benchmarking} with a fraction of the parameter count.

\begingroup
\setlength{\tabcolsep}{6pt}
\renewcommand{\arraystretch}{1.25}
\begin{table}[t]
  \centering
  \footnotesize
  \caption{Performance comparison on ARC-AGI benchmark. \textbf{Loop-ViT} demonstrates superior parameter efficiency, with a modest 11.2M parameters outperforming a 73M-parameter ensemble. Best results are \textbf{bold}, the second results are \underline{underlined}.}
  \label{tab:arc_results}
  \begin{tabular}{lccc}
    \toprule
    \textbf{Model} & \textbf{\#Params} & \textbf{ARC-AGI-1} & \textbf{ARC-AGI-2} \\
    \midrule
    
    \rowcolor{blue!10}
    \multicolumn{4}{c}{\textit{Large Language Models (LLMs)}} \\
    
    \rowcolor{gray!8}
    \textcolor{gray}{Deepseek R1~\cite{guo2025deepseek}} & \textcolor{gray}{671B} & \textcolor{gray}{15.8} & \textcolor{gray}{1.3} \\
    \rowcolor{gray!8}
    \textcolor{gray}{Claude 3.7 8k~\cite{arcprize2025arcagi_benchmarking}} & \textcolor{gray}{N/A} & \textcolor{gray}{21.2} & \textcolor{gray}{0.9} \\
    \rowcolor{gray!8}
    \textcolor{gray}{o3-mini-high~\cite{arcprize2025arcagi_benchmarking}} & \textcolor{gray}{N/A} & \textcolor{gray}{34.5} & \textcolor{gray}{3.0} \\
    \rowcolor{gray!8}
    \textcolor{gray}{GPT-5}~\cite{arcprize2025arcagi_benchmarking} & \textcolor{gray}{N/A} & \textcolor{gray}{44.0} & \textcolor{gray}{1.9} \\
    \rowcolor{gray!8}
    \textcolor{gray}{Grok-4-thinking~\cite{arcprize2025arcagi_benchmarking}} & \textcolor{gray}{1.7T} & \textcolor{gray}{66.7} & \textcolor{gray}{16.0} \\
    \rowcolor{gray!8}
    \textcolor{gray}{Bespoke (Grok-4)~\cite{berman2025highestscore}} & \textcolor{gray}{1.7T} & \textcolor{gray}{79.6} & \textcolor{gray}{29.4} \\
    
    \midrule
    \rowcolor{blue!10}
    \multicolumn{4}{c}{\textit{Recurrent Models}} \\
    
    HRM~\cite{wang2025hierarchicalreasoningmodel} & 27M & 40.3 & 5.0 \\
    TRM~\cite{jolicoeurmartineau2025morerecursivereasoningtiny} & 7M & 44.6 & 7.8 \\
    
    \midrule
    \rowcolor{blue!10}
    \multicolumn{4}{c}{\textit{Vision Models}} \\
    
    VARC~\cite{hu2025arc} & 18M & 54.5 & 8.3 \\
    VARC (ensemble)~\cite{hu2025arc} & 73M & 60.4 & 11.1 \\
    
    \midrule
    
    \textbf{Loop-ViT (Small)} & \textbf{3.8M} & 60.1 & 10.0 \\
    
    \textbf{Loop-ViT (Medium)} & \textbf{11.2M} & \underline{63.8} & \underline{11.5} \\
    \rowcolor{cyan!12}
    \textbf{Loop-ViT (Large)} & \textbf{18M} & \textbf{65.8} & \textbf{14.2} \\
    \midrule
    \rowcolor{blue!10}
    \multicolumn{4}{c}{\textit{human results}} \\    
    avg.human~\cite{legris2024harcrobustestimatehuman} & -- & 60.2 & -- \\
    best.human & -- & 98.0 & 100.0 \\
    \bottomrule
  \end{tabular}
\end{table}
\endgroup

Loop-ViT demonstrates a significant \textbf{Recurrence Dividend}. By recycling weights across iterations, Loop-ViT (Large) achieves $65.8\%$ on \mbox{ARC-1}, surpassing the $73$M VARC ensemble. This result establishes that scaling iterative computation (``Time'') is more effective for algorithmic induction than increasing raw layer counts (``Space'').

\subsection{Pillar 2: Ablation Study}
\label{subsec:ablations}
To evaluate the effectiveness of our design choices, we conduct a series of ablation experiments. We focus on two critical axes: (i) the structural trade-offs between parameter budget (space) and computational depth (time), and (ii) the necessity of spatial inductive biases in the recurrent core.

\begin{figure}[!t]
\centering
\includegraphics[width=1.00\linewidth]{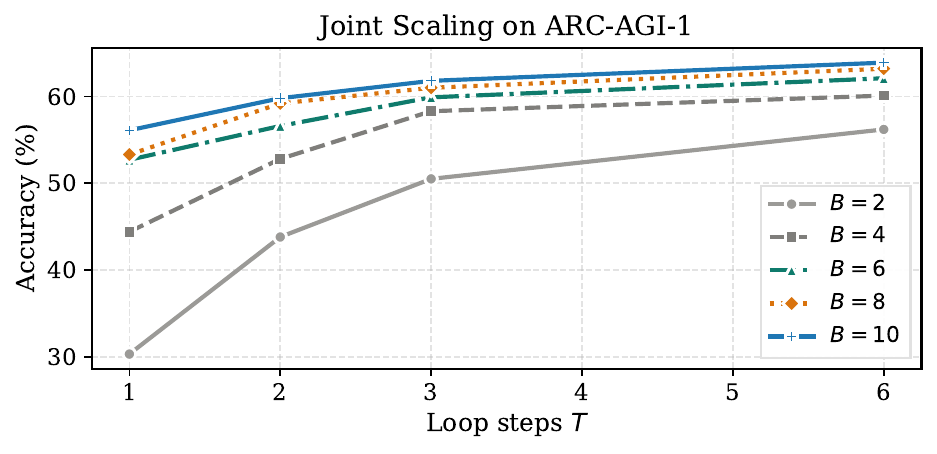}
\caption{\textbf{Joint scaling of core block depth ($B$) and loop steps ($T$) on \mbox{ARC-AGI-1}.} We vary the number of layers in the recurrent core ($B$) and the number of unrolled iterations ($T$). Each line represents a fixed core depth. The performance multiplier provided by recurrence is most evident in lower-capacity core models (\textit{e.g.}, $B=2$), where increasing $T$ from 1 to 6 yields a massive performance leap. Performance continues to scale with $T$ even for deeper cores, demonstrating that computational time can effectively compensate for limited parameter space.
}
\label{fig:joint_scaling}
\end{figure}

\begin{figure}[!t]
\centering
\includegraphics[width=1.00\linewidth]{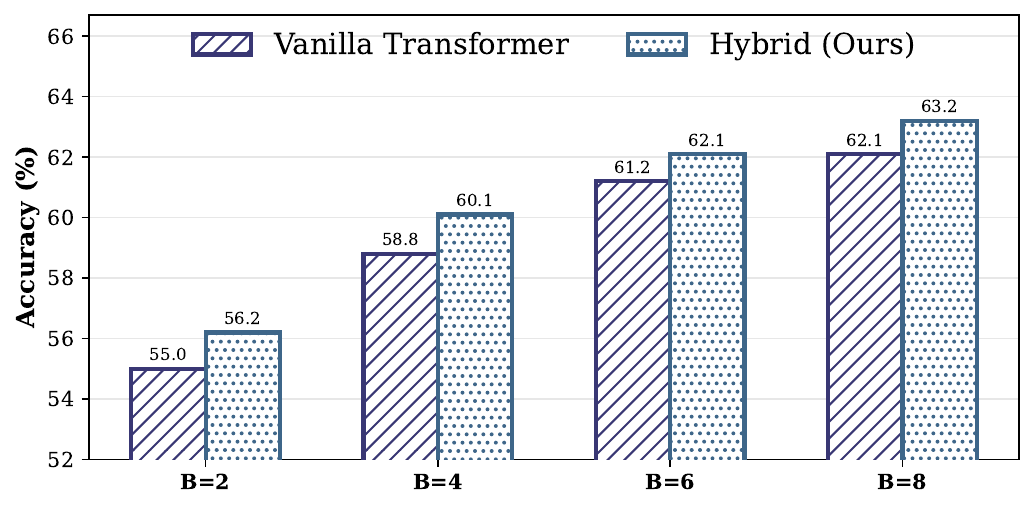}
\caption{\textbf{Ablation of Inductive Bias: Hybrid vs. Vanilla Core} across different core block depths ($B$) on \mbox{ARC-AGI-1}. We compare our Hybrid architecture (incorporating depth-wise convolutions in the FFN) against a standard Vanilla Transformer core. The Hybrid core consistently maintains a significant accuracy gap over the Vanilla baseline across all depths. This persistent advantage indicates that injecting local spatial priors is essential for grounding abstract reasoning in the image domain, and this requirement is not diminished by simply increasing model depth.
}
\label{fig:attention_archi}
\end{figure}

\paragraph{Space-Time Joint Scaling.}
\cref{fig:joint_scaling} explores the trade-off between core block depth ($B$) and the number of unrolled loop steps ($T$). The diverging trajectories reveal two regimes: (i) For low-capacity cores ($B=2$), increasing $T$ yields the most significant gains as weight-tied recurrence emulates the expressive depth of larger models; (ii) For high-capacity cores ($B=10$), performance continues to scale with $T$ up to the computational limit, reaching a peak of $63.9\%$.
\paragraph{Inductive Bias: Hybrid vs. Vanilla.}
The Hybrid Block (DW-Conv + MHSA) is compared against a standard Vanilla Transformer across core depths. As shown in \cref{fig:attention_archi}, the Hybrid architecture maintains a consistent lead. This suggests that local spatial priors are a fundamental requirement for grounding abstract reasoning in grid-based visual domains, regardless of model depth.

\begin{figure}[!t]
\centering
\includegraphics[width=1.00\linewidth]{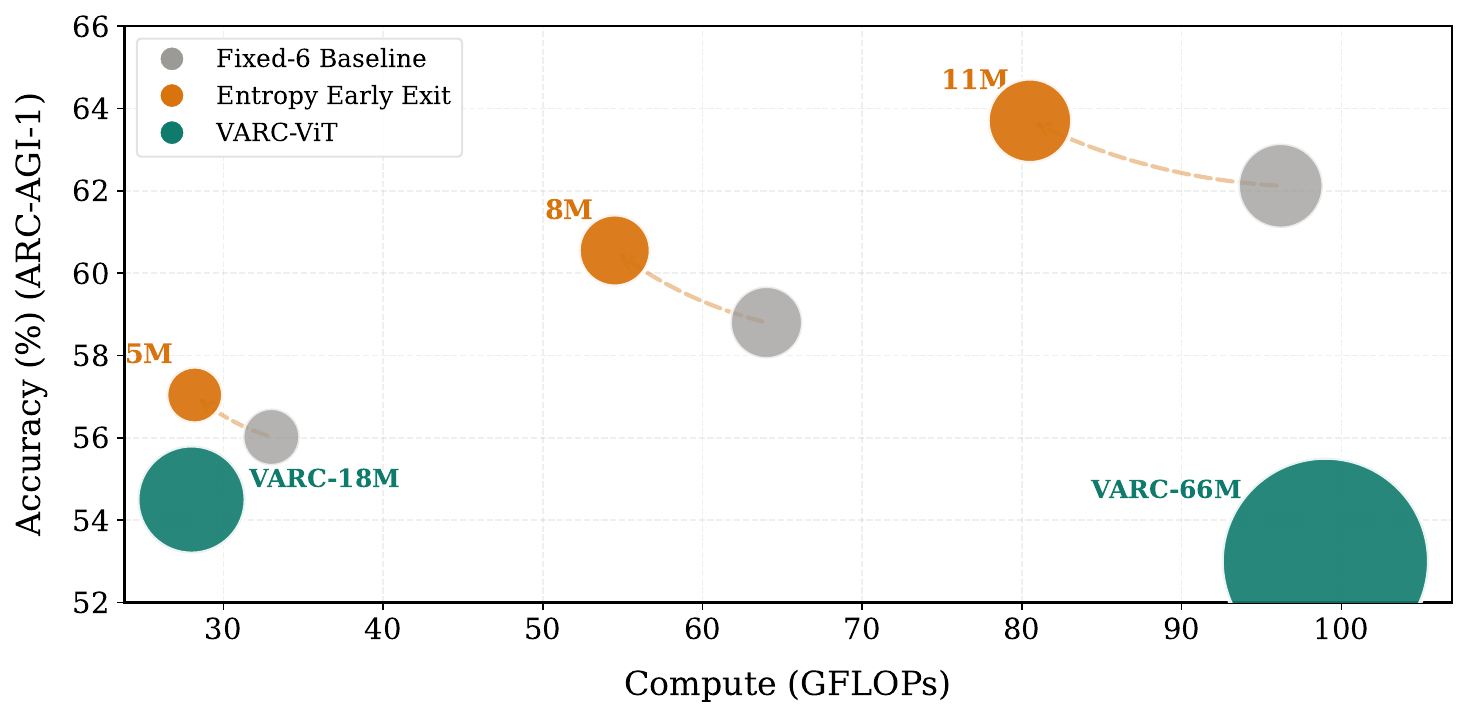}
\caption{\textbf{Accuracy--Compute--Params comparisons.} The horizontal axis is total inference compute, the vertical axis is Accuracy, and the circle radius corresponds to model Parameters. For Loop-ViT, GFLOPs accounts for unrolled recurrence and is computed using the \textit{average executed steps} under entropy-based halting. Our Entropy Early Exit strategy (orange) consistently surpasses the fixed-step baselines (grey) across all model scales ($B$=2, 4, 6), establishing a stronger accuracy--compute Pareto frontier. Feed-forward VARC baselines are included for comparison.}
\label{fig:abla_exit}
\end{figure}

\begin{figure}[!t]
\centering
\includegraphics[width=1.00\linewidth]{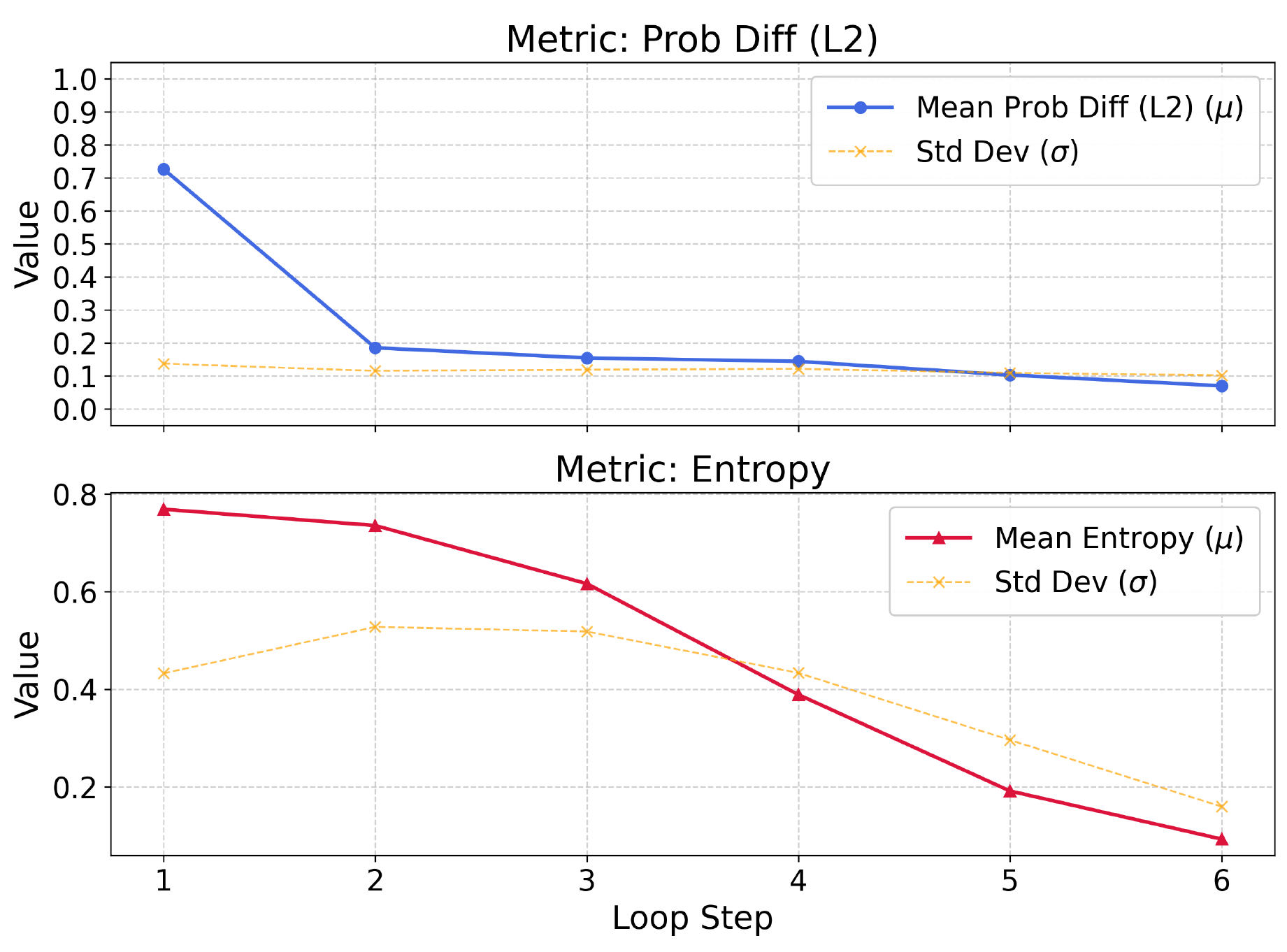}
\caption{\textbf{Quantitative Diagnostics of Recurrent Convergence.} We monitor the evolution of (top) the $L_2$-normalized difference $\delta_t$ and (bottom) the average pixel-wise Shannon entropy $\mathcal{H}_t$ across Loop-ViT iterations. Solid lines and shaded regions represent the mean and variance across the validation set, respectively. The synchronized decay of both prediction volatility and information uncertainty confirms that the model's internal state adheres to a stable trajectory toward a deterministic logical attractor, empirically validating our dynamic exit criterion.}
\label{fig:prob_diff_curve}
\end{figure}

\begin{figure}[t]
\centering
\includegraphics[width=1.00\linewidth]{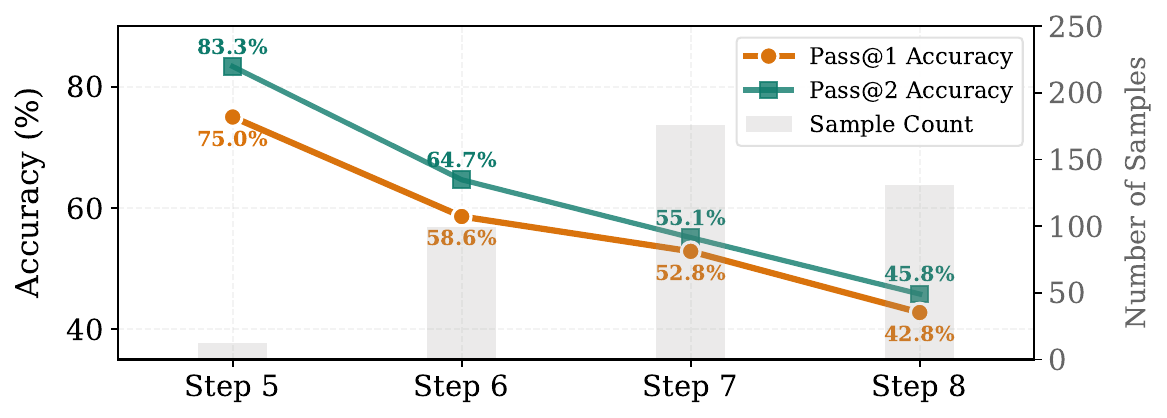}
\vspace{-1em}
\caption{\textbf{Efficiency vs. Task Difficulty Analysis.} Using the Loop-ViT variant ($B=2$), we stratify the test set by the number of inference steps Loop-ViT requires before exiting. ``Early Exit" (Step 5) samples achieve significantly higher accuracy (83.33\%) compared to those requiring the full depth (Step 8, 45.80\%). This confirms that the dynamic exit mechanism successfully identifies and solves simpler instances with minimal compute, while allocating more resources to harder tasks.}
\label{fig:efficiency}
\vspace{-1em}
\end{figure}

\begin{figure}[!t]
\centering
\includegraphics[width=1.00\linewidth]{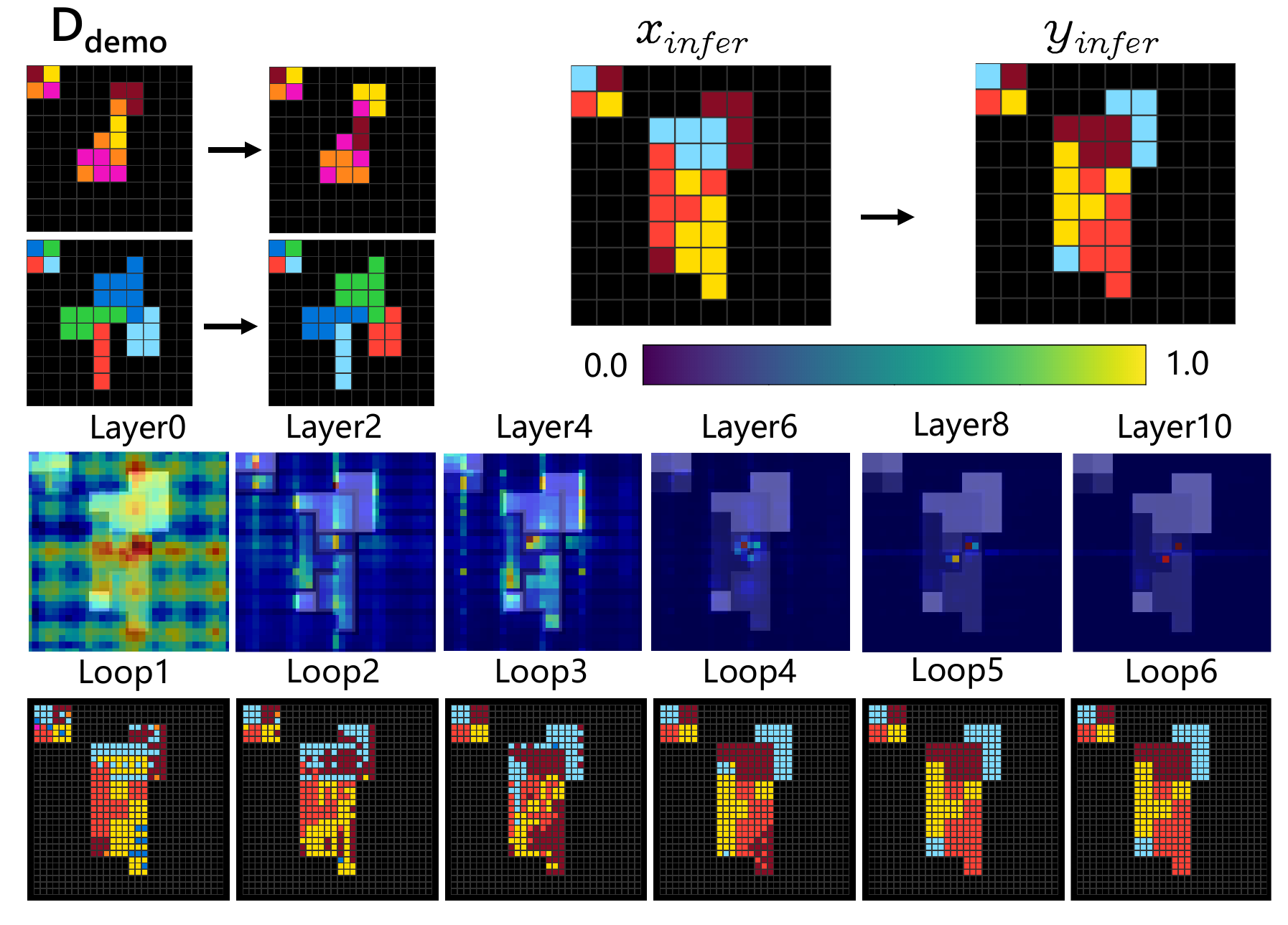}
\caption{\textbf{Evolution of Attention Patterns Across Processing Steps.} 
We visualize the average self-attention maps across Loop-ViT's recurrent steps. 
Early steps exhibit broad attention that analyzes the full input context. 
Later steps develop focused, sparse patterns that precisely track the algorithmic operations needed to solve the ARC task. 
This shift from global scanning to localized execution mirrors human reasoning strategies.
}
\label{fig:attention_evolution}
\end{figure}

\paragraph{Impact of Dynamic Exit.}
The effectiveness of adaptive halting is evaluated by comparing our dynamic-step model (constrained to $T \in [4, 8]$) against a fixed $6$-step baseline. As shown in \cref{fig:abla_exit}, the Dynamic Exit mechanism achieves higher accuracy while using lower \textit{average} inference compute on \mbox{ARC-AGI-1}. This is critical for weight-tied recurrent models, where inference compute scales approximately linearly with the number of executed iterations.

\paragraph{Efficiency vs. Difficulty.} 
We further analyze the correlation between inference steps and task difficulty using the Loop-ViT variant with $B=2$ (approximately 5M parameters) in \cref{fig:efficiency}. The results demonstrate a clear trend: samples that exit early (\textit{e.g.}, at Step 5) yield high accuracy ($83.33\%$), whereas those requiring more iterations (Step 8) are inherently more challenging ($45.80\%$). This validates that our entropy-based stopping criterion serves as an effective proxy for solution confidence, enabling the model to ``fast-track'' easier problems (Step 5) while instinctively reserving deeper computation for complex reasoning tasks (Step 8). This behavior mirrors human cognitive resource allocation, spending more time only when necessary.

\subsection{Mechanistic Insights}
\label{subsec:mechanistic}
Beyond quantitative performance, we conduct a qualitative analysis to understand the internal refinement process of Loop-ViT. By visualizing prediction crystallization and attention evolution, we aim to uncover how the recurrent state converges toward logically consistent solutions.

\paragraph{Prediction Crystallization.}
As illustrated in \cref{fig:overall}, Loop-ViT's predictions undergo a systematic ``crystallization'' process. Quantitative analysis in \cref{fig:prob_diff_curve} shows a synchronized decay in both prediction volatility ($L_2$ difference) and uncertainty (entropy). The $L_2$ difference drops precipitously in early iterations, suggesting a rapid commitment to the task geometry. The steady reduction in mean entropy $\mathcal{H}_t$ indicates the resolution of logical ambiguities.

\paragraph{Step-wise Attention Dynamics.}
We visualize the evolution of self-attention patterns across the loop in \cref{fig:attention_evolution}. In earlier steps, the attention matrices are relatively dense, reflecting a stage of \textit{Global Scanning} where the model integrates information from task demonstrators. As processing progresses, the attention shifts toward highly sparse and localized patterns. These later steps focus precisely on the grid transitions required for the predicted rule, corresponding to a stage of \textit{Local Execution}. This transition from exploratory to focused attention mirrors the deliberative strategies observed in human visual reasoning.
\section{Conclusion}
\label{conclu}
In this work, we introduced Loop-ViT, a recurrent vision architecture that challenges the paradigm of purely feed-forward visual reasoning. By decoupling reasoning depth from model capacity, we demonstrated that iterative computation is a more effective scaling axis than parameter width for abstract induction. 
Our design rests on two complementary pillars: a weight-tied Hybrid Block that aligns architectural inductive bias with the cellular nature of ARC transformations, and a Dynamic Exit mechanism driven by predictive entropy that enables the model to actively crystallize its latent state. 
Our results show that this simple approach significantly outperforms larger feed-forward baselines. We hope Loop-ViT serves as a strong baseline for future research on more complex reasoning tasks.

\small
\bibliographystyle{ieeenat_fullname}
\bibliography{reference}

\appendix
\onecolumn

\end{document}